\documentclass[12pt,pdftex]{l4dc2021} 

\title[Estimating Disentangled Belief about Hidden State and Hidden Task for Meta-RL]{Estimating Disentangled Belief about Hidden State and Hidden Task for Meta-Reinforcement Learning}
\usepackage{times}

\usepackage{amsmath,amsfonts,bm}

\def\eqref#1{equation~\ref{#1}}

\def\1{\bm{1}}

\DeclareMathAlphabet{\mathsfit}{\encodingdefault}{\sfdefault}{m}{sl}
\SetMathAlphabet{\mathsfit}{bold}{\encodingdefault}{\sfdefault}{bx}{n}

\DeclareMathOperator*{\argmax}{arg\,max}

\usepackage{mathtools}

\usepackage{url}
\usepackage{booktabs}       
\makeatletter
\newcommand{\figcaption}[1]{\def\@captype{figure}\caption{#1}}
\newcommand{\tblcaption}[1]{\def\@captype{table}\caption{#1}}
\makeatother

\author{\Name{Kei Akuzawa} \Email{akuzawa-kei@weblab.t.u-tokyo.ac.jp}\\
 \Name{Yusuke Iwasawa} \Email{iwasawa@weblab.t.u-tokyo.ac.jp}\\
 \Name{Yutaka Matsuo} \Email{matsuo@weblab.t.u-tokyo.ac.jp}\\
 \addr Graduate School of Engineering, The University of Tokyo, Japan}

\begin{document}

\maketitle

\begin{abstract}%


There is considerable interest in designing meta-reinforcement learning (meta-RL) algorithms, which enable autonomous agents to adapt new tasks from small amount of experience.
In meta-RL, the specification (such as reward function) of current task is hidden from the agent.
In addition, states are hidden within each task owing to sensor noise or limitations in realistic environments.
Therefore, the meta-RL agent faces the challenge of specifying both the hidden task and states based on small amount of experience.
To address this, we propose estimating disentangled belief about task and states, leveraging an inductive bias that the task and states can be regarded as global and local features of each task.
Specifically, we train a hierarchical state-space model (HSSM) parameterized by deep neural networks as an environment model, whose global and local latent variables correspond to task and states, respectively.
Because the HSSM does not allow analytical computation of posterior distribution, i.e., belief, we employ amortized inference to approximate it.
After the belief is obtained, we can augment observations of a model-free policy with the belief to efficiently train the policy.
Moreover, because task and state information are factorized and interpretable, the downstream policy training is facilitated compared with the prior methods that did not consider the hierarchical nature.
Empirical validations on a GridWorld environment confirm that the HSSM can separate the hidden task and states information.
Then, we compare the meta-RL agent with the HSSM to prior meta-RL methods in MuJoCo environments, and confirm that our agent requires less training data and reaches higher final performance.

\end{abstract}

\begin{keywords}%
  Meta-reinforcement learning, Partially observable Markov decision process, State space models, Amortized inference, Disentanglement
\end{keywords}

\section{Introduction}

The combination of reinforcement learning (RL) with deep learning has led to the rapid progress in difficult sequential decision making problems with high-dimensional observations \citep{mnih2015humanlevel}.
However, because conventional deep-RL methods learn a separate policy per task, it can lead to computationally intensive learning, requiring millions of interactions with one task.
Fortunately, many of the tasks tasks that we would like our agents to solve share common structure.
For example, in navigation tasks, an agent needs abilities to explore its surroundings, localize its location, and accurately map the environment, regardless of where the goal exists (that is, whatever the task is).
Meta-RL \citep{Duan2016RL2FR,finn2017maml} is a promising approach that exploits this structure to learn new tasks more quickly.
By training a policy using large quantities of experience collected across a distribution of tasks, it can quickly adapt to new tasks given a small amount of experience.

\begin{figure}[tbp]
  \begin{center}
   \includegraphics[width=140mm]{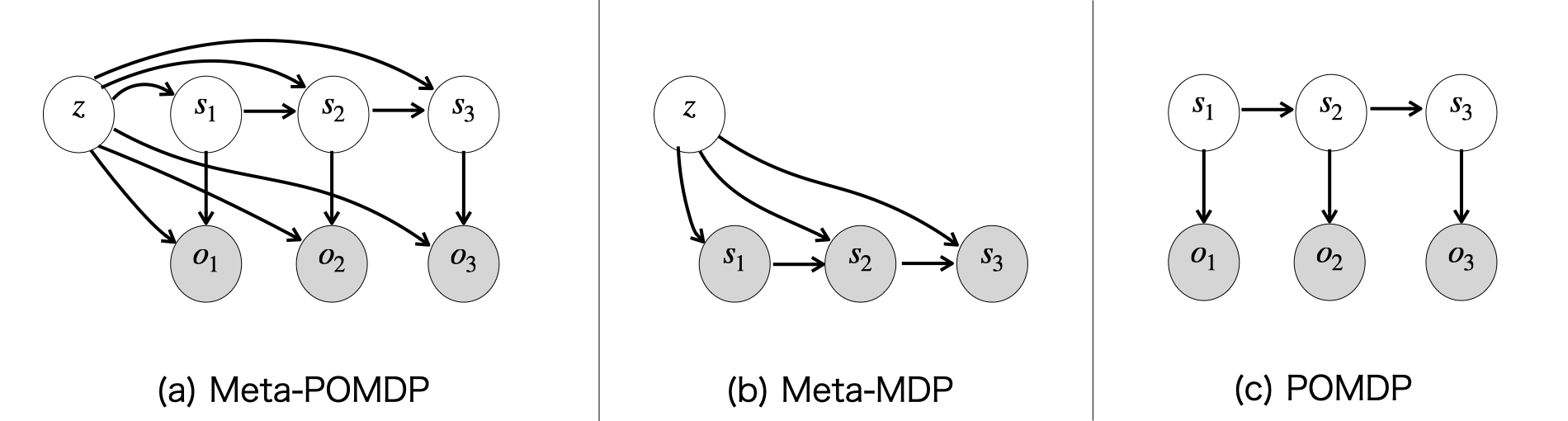}
  \end{center}
  \vskip -0.2in
  \caption{
    Comparison of Meta-POMDP, Meta-MDP, and POMDP.
    Note that the reward $r_t$ and action $a_t$ are omitted for simplicity.
    (a) In meta-POMDP, the state $s_t$ and task information $z$ are hidden from the agent.
    (b) In meta-MDP, only the task information $z$ is hidden.
    (c) In (single-task) POMDP, only  the state $s_t$ is hidden.
  }
  \label{fig:models}
 \vskip -0.3in
\end{figure}

To efficiently train the policy and enable fast adaptation in realistic environments, estimating both hidden task $z$ and states $s_t$ is significant.
First, meta-RL can be interpreted as a partially observable multi-task RL problem in which task specification (such as reward function and transition probability) is hidden from the agent \citep{humplik2019meta,Zintgraf2020VariBAD}.
Moreover, in realistic settings, sensor noise or sensor limitations may limit the agent's perceptual abilities, regardless of what the task is \citep{duan2016benchmarkRL,igl2018dvrl}.
Therefore, both $z$ and $s_t$ can be unobservable from the agent, that is, each task can be partially observable Markov decision process (POMDP, Figure \ref{fig:models}-(c)).
We refer to such an environment as meta-POMDP (Figure \ref{fig:models}-(a)) and distinguish it from meta-MDP (Figure \ref{fig:models}-(b)), in which states $s_t$ are observable within each task.
In meta-POMDP, it is not guaranteed that prior methods that assume meta-MDP structure \citep{rakelly2019pearl,Zintgraf2020VariBAD} will work well.
In addition, while some methods \citep{Duan2016RL2FR,zhao2020meld} were shown to work reasonably well in meta-POMDP, they did not distinguish $z$ and $s_t$ and treated them as a single ``hidden state.''
However, incorporating an inductive bias that $z$ and $s_t$ have hierarchical structure of Figure \ref{fig:models}-(a) may facilitate training and adaptation of the policy.

To support meta-RL in a realistic setting, we propose a natural and effective method for estimating disentangled belief states about $z$ and $s_t$, leveraging an inductive bias that the task and states can be regarded as global and local features of each task, respectively.
Specifically, we train a hierarchical state space model (HSSM) parameterized by deep neural networks (DNNs), which has the same data generating process as shown in Figure \ref{fig:models}-(a).
That is, our HSSM has a global latent variable $\tilde{z}$ and a local latent variable $\tilde{s}_t$, which correspond to $z$ and $s_t$, respectively.
Then, its posterior distribution for $\tilde{z}$ and $\tilde{s}_t$ approximates the belief, which summarizes past experience regarding the current task and states.
Although the posterior distribution cannot be analytically computed, we can approximate it using amortized inference \citep{kingma2013auto}.
After the belief is obtained, we can augment observations of a model-free policy with the belief to efficiently train the policy, as performed in prior studies \citep{Zintgraf2020VariBAD,zhao2020meld}.
However, because task and state information are encoded into separate latent variables, which are factorized and interpretable, the training and adaptation of the policy are facilitated compared with the prior methods.

In the meta-POMDP setting, (i) we experimentally confirm that our HSSM can disentangle hidden task information, using a GridWorld navigation environment.
Then, using the GridWorld and standard benchmark MuJoCo environments \citep{todorov2012mujoco}, we compare our method with prior meta-RL methods in terms of adaptation ability and training efficiency, following \cite{rakelly2019pearl}.
The results show that, (ii) the proposed method requires less training data and reaches higher final performance compared with the prior methods.
These experiments support our claim that 
the disentangled belief is a key ingredient for improving training efficiency and adaptation ability in the realistic meta-RL setting where both task and states are hidden.
Therefore, it is potentially beneficial for developing general-purpose robots that can solve various tasks with few trials.

\section{Meta-POMDP setting} \label{sec:preliminary}

Meta-POMDP is defined by task distribution $p(z)$ and a tuple $(S, O, A, R, T, T_{0}, F, \gamma, H^+)$, which depends on the realization of task $z \in Z$ (such as, a goal position or natural language instruction).
Here, $s_t \in S$, $o_t \in O$, $a_t \in A$, $\gamma$, and $H^+$ denote state, observation, action, discount factor, and episode length, respectively.
In addition, $R = p(r_t|s_t, s_{t-1}, a_{t-1}; z)$, $T = p(s_t|s_{t-1}, a_{t-1}; z)$, $T_{0} = p(x_0 | z)$, and $F = p(o_t|s_t; z)$ denote distribution of reward, transition, initial state, and observation, respectively.
Because these distributions can vary across tasks, they are conditioned on task $z$.
However, the agent cannot observe task $z$ as well as state $s_t$.
Moreover, meta-POMDP assumes the realistic environments where $S$, $Z$ $R$, $T$, $T_{0}$, and $F$ are also unknown.
This assumption distinguishes meta-POMDP from Bayes-adaptive POMDP \citep{stephane2008bapomdp}.
Although Bayes-adaptive POMDP assumes almost the same data generating process as meta-POMDP, it considers relatively small environments where $S$ and $O$ (state and observation spaces) are finite and known.

Because we adapt the standard meta-RL setting, a policy is first trained with experience collected across a distribution of tasks.
Then, the trained policy is evaluated on whether it can obtain higher rewards within the first $N$ POMDP episodes or at the $N$-th (final) episode.
Here, $N$ POMDP episodes with length $H$ are sampled from one meta-POMDP episode, that is, $H^+ = N \times H$.
When a single POMDP episode ends, the agent's state is reset based on the initial probability $T_0$, but task $z$ remains fixed until one meta-POMDP episode ends.

\section{Proposed method} \label{sec:proposed}

\subsection{Optimal policy in meta-POMDP with belief state} \label{sec:proposed:belief}

The solution for a meta-POMDP is a policy $\pi^* (s_t|\tau_{0:t})$ that maximizes discounted returns, i.e., $\pi^* = \argmax_\pi \mathbb{E}_{p_\pi}[\sum_{t=0}^{H^+} \gamma^t r_t]$.
However, using the entire trajectory $\tau_{0:t} = (o_{0:t}, a_{0:t-1}, r_{0:t-1})$ as an input is difficult to handle.
Fortunately, meta-POMDP can be regarded as a special case of POMDP by defining a new state $v_t^* = [s_t, z]$, i.e., regarding task as a type of hidden state.
This formulation enables the exploitation of \textit{belief states}, which has been well-studied in the POMDP literature \citep{leslie1998pomdp}.
Belief $b_t(v)$ is a sufficient statistic for optimal $a_t$ in the sense that there exists a policy $\pi (a_t|b_t)$ that satisfies $\pi^* (a_t|\tau_{0:t}) = \pi (a_t|b_t)$.
Because the belief can be represented in a lower dimension than $\tau_{0:t}$, the policy $\pi (a_t|b_t)$ can be trained efficiently.

Then, to efficiently train a policy in a meta-POMDP environment, the critical challenge is the estimation of the belief state.
The simplest way might be using the posterior distribution of the true environment model $b_t(v)=p(v_t^* = v|\tau_{0:t})$, which is known to satisfy the property of belief state \citep{leslie1998pomdp,Zintgraf2020VariBAD}.
In a small environment (Bayes-adaptive POMDP), the analytical method for calculating the posterior distribution is provided by \cite{stephane2008bapomdp}.
However, we consider a more realistic situation where the true environment model is unknown and the observation and state spaces could be infinite and unknown.
Therefore, we train an environment model to estimate the belief state, which is described in the next section

\subsection{Estimating belief state with hierarchical state space model} \label{sec:proposed:vae}

Because we assume that the true environment model is unknown, we propose learning the environment model of Figure \ref{fig:models}-(a) from data, and then, using posterior distribution of the model as the estimate of the belief.
To address this, we first parameterize the models of reward, transition, and observation with a DNN, and then, define the joint distribution of those models as a HSSM.
In designing the HSSM, we consider a situation where $S$ and $Z$ are unknown.
Then, we heuristically introduce alternative Euclidean spaces $\tilde{S}$ and $\tilde{Z}$ with $d_{\tilde{s}}$ and $d_{\tilde{z}}$ dimensions, respectively.
Using the $\tilde{S}$, $\tilde{Z}$, and model parameter $\theta$, we design reward model $R_\theta$, transition model $T_\theta$, initial state probability $T_{\theta, 0}$, observation model $F_\theta$, and task distribution $\tilde{Z}_\theta$ as follows:
\begin{align}
    R_\theta &= p_\theta (r_t | \tilde{s}_t, \tilde{s}_{t-1}, a_{t-1}, \tilde{z}) = N (f_{r, \mu} (\tilde{s}_t, \tilde{z}), f_{r, \sigma} (\tilde{s}_t, \tilde{z})), \\
    T_\theta &= p_\theta (\tilde{s}_t | a_{t-1}, \tilde{s}_{t-1}, z) = N (f_{\tilde{s}, \mu} (a_{t-1}, \tilde{s}_{t-1}, \tilde{z}), f_{\tilde{s}, \sigma} (a_{t-1}, \tilde{s}_{t-1}, \tilde{z})), \\
    F_\theta &= p_\theta (o_t | \tilde{s}_t, \tilde{z}) = N (f_{o, \mu} (\tilde{s}_t, \tilde{z}), f_{o, \sigma} (\tilde{s}_t, \tilde{z})), \\
    T_{\theta, 0} &= p_\theta (\tilde{s}_0) = N (\bf{0}, \bf{I}), \\
    \tilde{Z}_\theta &= p_\theta (\tilde{z}) = N (\bf{0}, \bf{I}),
\end{align}
where $\tilde{s_t} \in \tilde{S}$ and $\tilde{z} \in \tilde{Z}$.
Here, each $f$ is a feedforward neural network with parameter $\theta$, and represents normal distribution using a reparameterization trick \citep{kingma2013auto}.
In addition, $T_\theta$, $R_\theta$, and $F_\theta$ are conditional on $\tilde{z}$, i.e., they may change depending on the task.
Note that, for simplicity, we assume that $R_\theta$ only depends on $\tilde{s}_t$ and $\tilde{z}$, i.e., $p_\theta (r_t | \tilde{s}_t, \tilde{s}_{t-1}, a_{t-1}, \tilde{z}) = p_\theta (r_t | \tilde{s}_t, \tilde{z})$.
In addition, we parameterize $F_\theta$ as normal distribution here, but other distributions such as Bernoulli distribution can be appropriate depending on the environment.

Using these distributions, the joint distribution of the models can be expressed as follows:
\begin{align}
    p_\theta(\tau_{0:T}|a_{0:T-1}) =& \int \int p_\theta(\tilde{z}) p_\theta(\tilde{s}_0) \\
                                    & \sum_{t=1}^T p_\theta(\tilde{s}_t|\tilde{s}_{t-1}, a_{t-1}, \tilde{z}) p_\theta(r_t|\tilde{s}_t, \tilde{s}_{t-1}, a_{t-1}, \tilde{z})
                                     p_\theta(o_t|\tilde{s}_t, \tilde{z}) d\tilde{z} d\tilde{s}_{0:T}.
\end{align}
Then, when the generative model $p_\theta(\tau_{0:T}|a_{0:T-1})$ is trained and approximates the true environment well, the posterior distribution of the model $\hat{b}_t \coloneqq p_\theta (\tilde{s}_t, \tilde{z} | \tau_{0:t})$ can be used as the estimate of the belief $b_t$.
However, this approach has two challenges:
(i) the posterior distribution $p_\theta (\tilde{s}_t, \tilde{z}|\tau_{0:t})$ cannot be calculated analytically, and then (ii) the likelihood of the model is intractable.

To address these issues, we employed amortized inference that performs approximate inference on the two latent variables.
We prepare two encoders $q_\phi(\tilde{z}|\tau_{0:t})$ and $q_\phi(\tilde{s}_t|\tau_{0:t})$ parameterized by $\phi$, which represent normal distribution using a reparameterization trick.
Specifically, the encoders first compress trajectories $\tau_{0:t}$ into hidden state $h_t$ with recurrent neural networks (RNNs).
Then, they output the mean and variance from $h_t$ using feedforward neural networks $g$ as follows:
\begin{align}
    q_\phi (\tilde{s}_t | \tau_{0:t}) = N (\mu_{\tilde{s}} = g_{\tilde{s}, \mu} (h_t), \sigma_{\tilde{s}} = g_{\tilde{s}, \sigma} (h_t)), \\
    q_\phi (\tilde{z} | \tau_{0:t}) = N (\mu_{\tilde{z}} = g_{\tilde{z}, \mu} (h_t), \sigma_{\tilde{z}} = g_{\tilde{z}, \sigma} (h_t)).
\end{align}
Using $q_\phi$, the evidence lower bound (ELBO) of the likelihood can be calculated as follows:
\begin{align}
    \mathcal{L} =& \mathbb{E}_{q_\phi(\tilde{z}|\tau_{0:T}) \Pi_{t=0}^T q_\phi(\tilde{s}_{t}|\tau_{0:t})} \sum_{t=0}^T \biggl([\log p_\theta(o_t|\tilde{s}_t, \tilde{z})
                                                                                                                              + \log p_\theta(r_t|\tilde{s}_{t}, \tilde{s}_{t-1}, a_{t-1}, \tilde{z})] \\
                     &- \mathrm{D_{KL}} \bigl[q_\phi(\tilde{s}_t|\tau_{0:t}) | p_\theta(\tilde{s}_t|\tilde{s}_{t-1}, a_{t-1}, \tilde{z})\bigr]
                      - \mathrm{D_{KL}} \bigl[q_\phi(\tilde{z}|\tau_{0:T}) | p_\theta(\tilde{z})\bigr] \biggr). \label{eq:elbo}
\end{align}
To optimize the ELBO, we approximate the expectation over $q_\phi(\tilde{z}|\tau_{0:T})$ and $q_\phi(\tilde{s}_{t}|\tau_{0:t})$ with Monte Carlo sampling, as performed in standard variational autoencoders (VAEs) \citep{kingma2013auto}.
Then, the reconstruction losses (the first and second terms) and the Kullback-Leibler (KL) divergences (the third and fourth terms) can be computed analytically; therefore, the environmental model can be trained with the gradient ascend to maximize the ELBO.

\begin{figure}[tbp]
  \begin{center}
   \includegraphics[width=140mm]{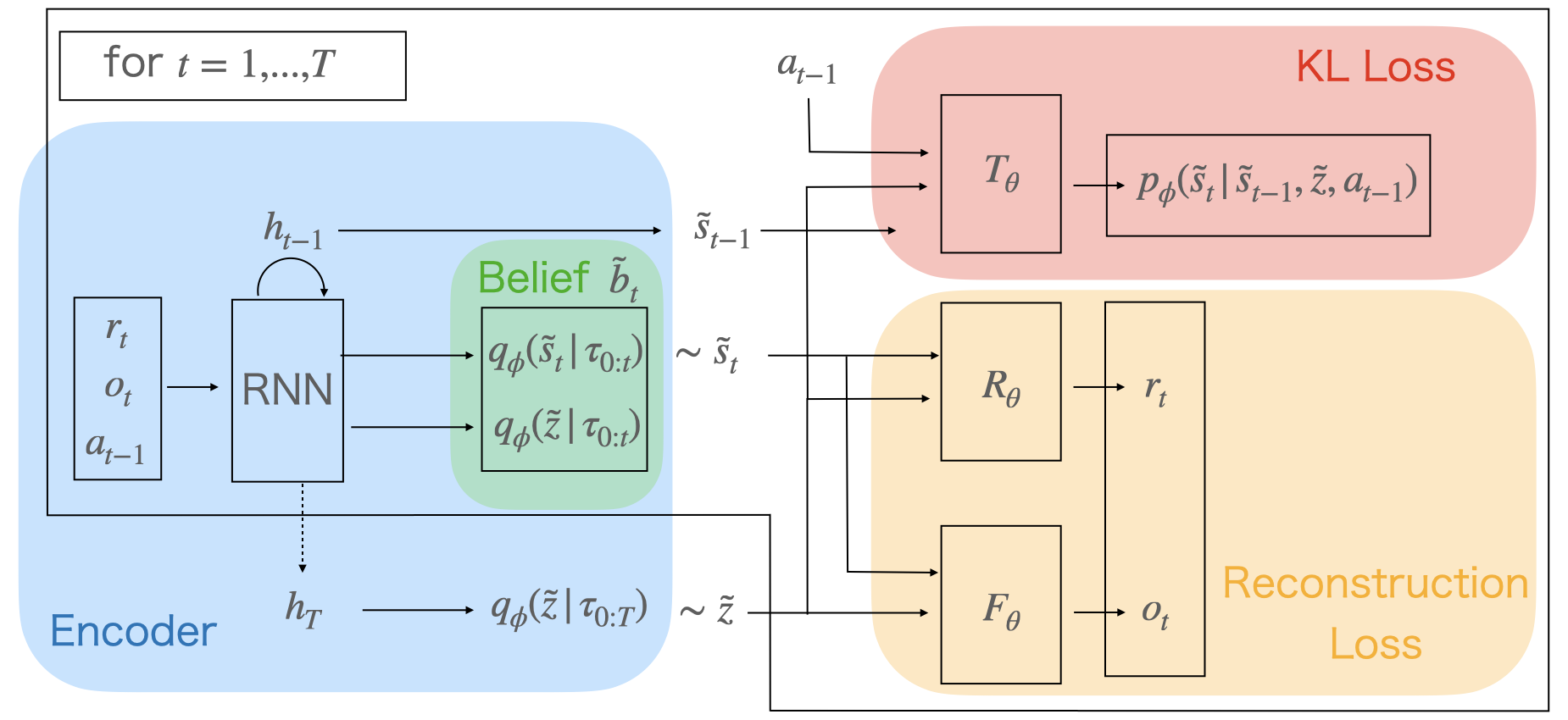}
  \end{center}
  \vskip -0.15in
  \caption{The proposed HSSM architecture.
  When training the HSSM, an RNN produces the posteriors over task $q_\phi(\tilde{z}|\tau_{0:T})$ and state $q_\phi(\tilde{s}_t|\tau_{0:t})$ by encoding the entire history of $\tilde{s}_t$, $a_t$, and $r_t$.
  The posteriors are trained with reconstruction and KL losses, along with transition $T_\theta$, reward $R_\theta$, and observation model $F_\theta$.
  When the agent acts in the environment, the observation is augmented by $\tilde{b}_t = q_\phi(\tilde{z}|\tau_{0:t})q_\phi(\tilde{s}_t|\tau_{0:t})$, which is calculated online.
  }
  \label{fig:architecture}
 \vskip -0.15in
\end{figure}

After we obtain the learned environment model, we can approximate the belief using the encoders.
First, as in standard VAEs, the encoder distribution $q_\phi(\tilde{s}_t, \tilde{z}|\tau_{0:T}) \coloneqq q_\phi(\tilde{s}_t|\tau_{0:t})q_\phi(\tilde{z}| \tau_{0:T})$ approximate the posterior distribution of the model $p_\theta(\tilde{s}_{t}, \tilde{z}|\tau_{0:T}) = p_\theta(\tilde{s}_{t}|\tau_{0:T}, \tilde{z})p_\theta(\tilde{z}| \tau_{0:T})$ because the ELBO and the likelihood have the following relationship:
\begin{align}
    \mathcal{L} =& \log p_\theta(\tau_{0:T} | a_{0:T}) - \mathrm{D_{KL}} [ q_\phi(\tilde{z}, \tilde{s}_{0:T}| \tau_{0:T}) | p_\theta(\tilde{z}, \tilde{s}_{0:T} | \tau_{0:T}) ].
\end{align}
Thus, the minimization of the ELBO not only increases the likelihood, but also minimizes the approximation error $\mathrm{D_{KL}} [ q_\phi(\tilde{z}, \tilde{s}_{0:T}| \tau_{0:T}) | p_\theta(\tilde{z}, \tilde{s}_{0:T} | \tau_{0:T}) ]$.
When the encoder well approximates $p_\theta(\tilde{s}_t, \tilde{z} | \tau_{0:t})$, it also approximates the belief, that is, true posterior $b_t = p(s_t, z | \tau_{0:t})$, as follows:
\begin{align}
    b_t \approx \hat{b}_t = p_\theta(\tilde{s}_t, \tilde{z}| \tau_{0:t}) \approx q_\phi(\tilde{s}_t|\tau_{0:t}) q_\phi(\tilde{z}|\tau_{0:t}) \eqqcolon \tilde{b}_t. \label{eq:belief_approx}
\end{align}
Here, $q_\phi(\tilde{s}_t|\tau_{0:t})$ and $q_\phi(\tilde{z}|\tau_{0:t})$ represent $d_{\tilde{s}}$- and $d_{\tilde{z}}$-dimensional normal distributions, respectively.
Then, in practice, we use $2 (d_{\tilde{s}} + d_{\tilde{z}})$-dimensional vector $\tilde{b}_t = [\mu_{\tilde{z}}, \sigma_{\tilde{z}}, \mu_{\tilde{s}}, \sigma_{\tilde{s}}]$ as the estimate of $b_t$.
The architectural diagram of the HSSM is illustrated in Figure \ref{fig:architecture}.

Finally, we note that $\tilde{b}_t$ has two approximation errors as shown in eq. \ref{eq:belief_approx}:
(i) the difference between model $p_\theta$ and true environment $p$, and (ii) the difference between model posterior $p_\theta$ and amortized posterior $q_\phi$.
This study does not consider the errors because the scope of this study is to propose a natural method for estimating belief state in the realistic meta-RL environment where both task and state are hidden.
However, it could be possible to reduce the errors by (i) using deeper neural networks \citep{vahdat2020NVAE} to increase the model capacity, or (ii) using normalizing flows such that the amortized posterior can express more flexible distribution \citep{kingma2016iaf}.

\subsection{Training policy with disentangled belief}

As explained in Section \ref{sec:proposed:belief}, $b_t$ is a sufficient statistic for the optimal $a_t$ as well as lower dimensional.
Then, we propose training a policy $\pi_{\psi}(a_t|\tilde{b}_t, o_t)$ parameterized by feedforward neural networks $\phi$, which receives $\tilde{b}_t$ as inputs.
Because this approach separates learning decision making (policy) from learning representation (HSSM), the policy can be efficiently trained with a model-free algorithm such as PPO \citep{schulman2017ppo} or SAC \citep{haarnoja18sac}.
Moreover, belief $\tilde{b}_t$ is also factorized and interpretable, that is, task and state information are hopefully contained in $\tilde{z}$ and $\tilde{s}$, respectively.
Such disentangled representation helps in efficient learning of the policy, compared with the method that does not distinguish the hidden task and states \citep{zhao2020meld} (further discussed in Section \ref{sec:related}).
We refer to this meta-RL method of ours as HSSM-agent (HSSMA).

Because the approach in which the belief state is inputted to a model-free policy has already succeeded in POMDP \citep{han2020variational}, meta-MDP \citep{Zintgraf2020VariBAD}, and meta-POMDP \citep{zhao2020meld} environments, we adapt some training techniques from them.
First, we optimize the policy and the model (HSSM) using different optimizers alternately, without backpropagating the RL loss through the encoder (i.e., end-to-end training).
Specifically, the update of the policy $\pi_{\psi}(a_t|\tilde{b}_t, o_t)$ originally depends on the encoder parameter $\phi$ because $\tilde{b_t}$ is calculated using $\phi$.
However, as \cite{Zintgraf2020VariBAD} reported, we found that the end-to-end training is typically unnecessary in practice.
Second, as in \cite{han2020variational}, we condition the policy with raw observations $o_t$ in order to stabilize the early stage of the training in which $\tilde{b}_t$ has not approximated $b_t$ well.

\subsection{Posterior collapse and $\beta$-VAE objective} \label{sec:proposed:beta}

In practice, we observe that the optimization of the HSSM sometimes suffers from posterior collapse.
The posterior collapse often occurs in hierarchical VAEs like our HSSM, in which the higher latent variable $\tilde{z}$ becomes uninformative as the data distribution is solely modeled by the lower latent variables \citep{hsu2017fhvae,maalo2019biva}.
To alleviate this problem, we employed $\beta$-VAE objective proposed by \cite{alemi2018fixing}, which regularizes mutual information between $\tilde{z}$ and $\tau_{0:T}$ to be large such that $\tilde{z}$ becomes informative.
Specifically, we modify the objective function (Eq. \ref{eq:elbo}) using a weighting parameter $\beta < 1$ as follows:
\begin{align}
    \mathcal{L_\beta} =& \mathbb{E}_{q_\phi(\tilde{z}|\tau_{0:T}) \Pi_{t=0}^T q_\phi(\tilde{s}_{t}|\tau_{0:T})} \sum_{t=0}^T \biggl([\log p_\theta(o_t|\tilde{s}_t, \tilde{z})
                      + \log p_\theta(r_t|\tilde{s}_{t}, \tilde{s}_{t-1}, a_{t-1}, \tilde{z})] \\
                     &- \mathrm{D_{KL}} \bigl[q_\phi(\tilde{s}_t|\tau_{0:T}) | p_\theta(\tilde{s}_t|\tilde{s}_{t-1}, a_{t-1},\tilde{z})\bigr]
                      - \beta \mathrm{D_{KL}} \bigl[q_\phi(\tilde{z}|\tau_{0:T}) | p_\theta(\tilde{z})\bigr] \biggr)
\end{align}
Note that, \cite{alemi2018fixing} use $\beta < 1$ to regularize mutual information to be large, although $\beta$-VAE was originally invented to encourage the independence of each dimension of $\tilde{z}$ with $\beta > 1$ by \cite{higgins2017beta}.
The effect of this weighting parameter is confirmed through our experiments.

\section{Related work} \label{sec:related}
Most relevant to this study is the belief-based approach for meta-RL.
An early method RL$^2$ \citep{Duan2016RL2FR} aggregated past experience using RNN and implicitly extracted the belief.
However, it requires simultaneously learning task representation and decision making using only reward signal, making the optimization difficult.
Following this, PEARL \citep{rakelly2019pearl} and VariBAD \citep{Zintgraf2020VariBAD} decoupled the learning of task representation from that of decision making.
However, they estimated the belief regarding only the task, assuming that the states could be observed.
In addition, because meta-POMDP is a special case of POMDP, the methods proposed in the POMDP literature \citep{igl2018dvrl,lee2019slac,han2020variational,gregor2019shaping} could in principle be applied to meta-POMDP.
Then, \cite{zhao2020meld} proposed MELD, which applied deep SSM of \cite{lee2019slac} to estimate the belief.
Unlike these methods, our method has both $\tilde{z}$ and $\tilde{s}_t$, which allows us to disentangle \citep{Bengio+2013_RLsurvey} the task and state information.
Disentangling factors of variation facilitates the learning of classifier and RL agent, as shown in the experiments of \cite{higgins2017beta,higgins2017darla} and ours.
In addition, it allows the reasoning about task-level uncertainty, which might facilitate active learning and safety-critical applications \citep{finn2018pmaml}, while we do not experiment with these in this paper.

The approaches, including ours and MELD, in which observations are compressed to improve the sample efficiency of a policy, are called state representation learning (SRL) \citep{Lesort2018StateRL}.
In the literature on SRL, some studies have proposed methods that consider a hierarchical property of time.
For example, \cite{boehmer13sfa,Jonschkowski2015prior} proposed extraction of features that are slowly varying, assuming that interesting features fluctuate slowly and continuously through time.
The proposed method is similar to these studies in incorporating the hierarchical property of time, but it focuses on the meta-POMDP environment where a single task is composed of time-invariant global features (task information) and time-variant local features (states).

From a technical perspective, our HSSM relates to the literature on the deep sequential latent variable models.
Including our study, many studies have proposed to disentangle time-invariant and time-variant features using the latent variable models and amortized inference, in the domains of text \citep{bowman2016generating}, image \citep{chen2017vlae}, movie \citep{hsieh2018video}, and speech \citep{hsu2017fhvae,yingzhen2018dsae}.
The architecture of our HSSM is inspired from \cite{yingzhen2018dsae}, although some details are different (e.g., our transition model is conditioned on the global latent variable).
In addition, these studies found that a global latent variable tends to be ignored owing to posterior collapse, which motivates us to employ $\beta$-VAE objective in Section \ref{sec:proposed:beta}.


Although this study assumes that task information is hidden in line with the meta-RL literature, many studies also consider the multi-task RL setting where task information is available.
For example, \cite{humplik2019meta} argued that auxiliary information about task (such as task ID) is sometimes available, and proposed to exploit them in estimating the belief.
In addition, many studies (e.g., \cite{jiang2019lang,chevalier-boisvert2018babyai}) have tried to use language instructions as task specifications that define the environment model (such as reward function).
While using such auxiliary information about the task can improve the performance of the agent, it can be expensive.
However, since the proposed model treats the task as latent variables, it could be extended to balance the performance and the labeling cost.
That is, our method can be extended to the semi-supervised setting in which the task information is occasionally available in a similar way to \cite{kingma2014ssvae}, which is also the potential merit of disentangling task information from states.


\section{Experiment} \label{sec:exp}

\subsection{Setting}
We design the experiments to answer the following questions:
(i) Can the proposed HSSM disentangle task information into the global latent variable $\tilde{z}$ in meta-POMDP environments?
(ii) Does the disentangled belief facilitate the adaptation and training of a meta-RL policy?

To address this, we compared {\bf HSSMA} with the following methods:
{\bf (i) RL$^2$} is a meta-RL method using a RNN policy.
It was shown to work reasonably well in meta-POMDP.
{\bf (ii) VariBAD} is a state-of-the-art meta-RL method, which outperformed RL$^2$ in a meta-MDP setting.
{\bf (iii) SSM-agent (SSMA)} is an ablation method, which is the same with HSSMA except for not having $\tilde{z}$.
In addition, SSMA is very similar to MELD, although their architectural details are different.
In all the methods, we train a PPO policy parameterized by feedforward neural networks with $1e+8$ frames.

We evaluated the methods using the following environments, following \cite{Zintgraf2020VariBAD}:
{\bf GridWorld:}
Here, the task is to reach a goal in a 5 × 5 gridworld.
The goal position is randomly chosen per task, and the agent gets a sparse reward signal: -0.1 on non-goal cells and +1 on the goal cell.
Therefore, reward function $R$ varies across tasks.
In addition, the observation of the agent is 3 x 3 cells around the agent.
Specifically, the agent only knows whether an adjacent cell is a wall or a floor.
The horizon within one POMDP episode is $H = 15$, and the agent is reset to the initial position after the $H=15$ steps.
Also, each task consists of four POMDP episodes, i.e., $N = 4$.
Furthermore, we employed two MuJoCo \citep{todorov2012mujoco} locomotion environments commonly used in the meta-RL literature.
{\bf HalfCheetahVel:}
Here, the agent has to run at different velocities per task, i.e., the agent gets a reward defined by the distance between its velocity and the target one.
Therefore, $R$ varies across tasks.
{\bf Walker2DRandParam:}
Here, the system parameters are randomized per task.
Then, transition probability $T$ varies across tasks.
In these environments, we set $H = 200$ and $N = 2$.
In addition, following \cite{duan2016benchmarkRL}, we make these MuJoCo environments partially observable by limiting sensors;
we restrict the observations to only provide positional information (including joint angles), excluding velocities.

HSSMA primarily has three hyperparameters: the dimension size of $\tilde{s}_t$ and $\tilde{z}$, ($d_{\tilde{s}}$ and $d_{\tilde{z}}$), and the $\beta$ value in $\beta$-VAE objective (Section \ref{sec:proposed:beta}).
In GridWorld, we set $d_{\tilde{s}}=5$ and $d_{\tilde{z}}=5$.
In addition, in MuJoCo environments, we set $d_{\tilde{s}}=64$ and $d_{\tilde{z}}=32$.
The baseline methods also have a dimension size of hidden state as a hyperparameter.
For fair comparison, we designed them to be equal to the sum of $d_{\tilde{z}}$ and $d_{\tilde{s}}$ in most cases.
However, the dimension size of RL$^2$ in GridWorld is set to $32$ in accordance with the experiment of \cite{Zintgraf2020VariBAD}.
Regarding $\beta$, we tested $(1, 1e-1, 1e-2)$ in GridWorld.
As noted later, HSSMA with $\beta=1e-1$ achieved better performance than that with $\beta=1$, so we set $\beta=1e-1$ in MuJoCo environments.
In the experiments, we reported that the scores averaged over three random seed trials for GridWorld, and five trials for MuJoCo


\subsection{Contribution of global latent variable on fast adaptation}

\begin{figure}[tbp]
    \def\@captype{table}
    \begin{minipage}{0.53\hsize}
        \tblcaption{Accuracies of logistic regression from latent variables to goal position (task information) in GridWorld.
        The $\uparrow$ and $\downarrow$ indicate that the purpose was to obtain high and low scores, respectively.
        }
        \scalebox{0.8}{
        \begin{tabular}{lllrlll}
        \toprule
        {} & Model & $\beta$ & $d_{\tilde{s}}$ & $\tilde{z}$-accuracy $\uparrow$ & $\tilde{s}$-accuracy $\downarrow$ & $\tilde{z}$-KL \\
        \midrule
         &  HSSM &  1e-0 &      5 &      44.00 &      47.00 &  2.12 \\
         &  HSSM &  1e-1 &      5 &      80.67 &     {\bf 33.50} &  5.48 \\
         &  HSSM &  1e-2 &      5 &     {\bf 90.67} &      44.17 &  12.3 \\
        \hline
         &   SSM &   N/A &      5 &        N/A &      48.00 &   N/A \\
         &   SSM &   N/A &     10 &        N/A &      50.17 &   N/A \\
        \bottomrule
        \end{tabular}

        }
        \label{tab:exp:rep:grid}
    \end{minipage}
    \def\@captype{table}
    \begin{minipage}{0.47\hsize}
        \centering
        \includegraphics[width=\columnwidth]{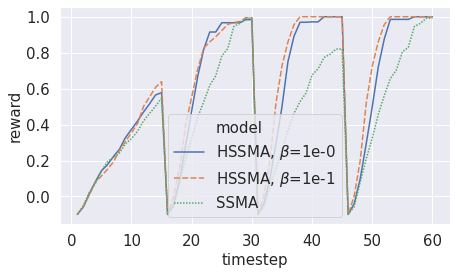}
        \vskip -0.2in
        \figcaption{A reward (y-axis) for each timestep (x-axis) within single task in GridWorld.
                    One POMDP episode is reset for each 15-timestep.
        }
        \label{fig:exp:grid-beta}
    \end{minipage}
    \vskip -0.25in
\end{figure}

We first evaluate whether the HSSM can disentangle task information into the global latent variable $\tilde{z}$, using GridWorld.
Specifically, we used the learned encoder $q_\phi (\tilde{z} | \tau_{0:T})$ to extract its mean $\mu_{\tilde{z}}$ from 1000 trajectories where $T = N \times H =60$.
At the same time, we prepared true task information $z$ for the trajectories.
Here, $z$ (goal position) is a 21-class discrete variable, which can be anywhere except around the starting cell at the bottom left.
Then, we split the pairs of $(\mu_{\tilde{z}}, z)$ into 800 training and 200 test samples, and evaluated the accuracy of logistic regression from $\mu_{\tilde{z}}$ to $z$ (denoted as $\tilde{z}$-accuracy).
Here, we chose to use the simple logistic classifier because it is likely that when the classifier perform well with a small number of samples, a downstream policy can be efficiently trained.
In addition, to confirm that $\tilde{z}$ and $\tilde{s}_t$ are disentangled, that is, $\tilde{s}_t$ has no task information, we performed classification from $\mu_{\tilde{s}}$ at time $T$ to $z$ (denoted as $\tilde{s}$-accuracy).
In addition, $\tilde{s}$-accuracy was also evaluated for SSM (the environment model of SSMA).
Furthermore, we evaluated the value of the KL term for $\tilde{z}$ (denoted as $\tilde{z}$-KL), which approximates the amount of information in $\tilde{z}$.

Table \ref{tab:exp:rep:grid} shows the results of the experiment.
The table shows that, (i) when the value of $\beta$ is extremely high, $\beta=1$, $\tilde{z}$-accuracy, and $\tilde{z}$-KL are low.
In contrast, when $\beta=1e-1$ or $1e-2$, $\tilde{z}$-accuracy increases.
This indicates that $\tilde{z}$ is ignored owing to posterior collapse when $\beta$ is extremely large, but we can alleviate the problem and obtain meaningful $\tilde{z}$ with appropriate $\beta$ values.
In addition, (ii) $\tilde{z}$-accuracy and $s$-accuracy of the HSSM are higher and lower than the $\tilde{s}$-accuracy of SSM, respectively.
This may be due to the fact that the baseline SSM do not explicitly separate the hidden task from the hidden state, so its latent representation is entangled.

Next, in GridWorld, we evaluate the adaptation abilities of HSSMA with various $\beta$ values and SSMA.
Figure \ref{fig:exp:grid-beta} shows a reward for each timestep within single task, consisting of four POMDP episodes.
This figure shows that although all the methods have roughly the same rewards for the first episode, HSSMA with $\beta=1e-1$ achieved higher rewards in the third to fourth episodes.
In other words, SSMA and HSSMA with $\beta=1$, which suffer from posterior collapse, have unstable and lower rewards in the later episodes.
This indicates that these methods are not appropriate for making use of knowledge about task based on past experience when facing a new episode because they do not explicitly retrieve task information.
Also, note that the returns for $\beta=1e-1$ and $\beta=1e-2$ are almost the same; therefore, we omitted the result for $\beta=1e-2$ here.

\subsection{Comparing meta-RL performance with prior methods}

\begin{figure}[t]
\begin{center}
    \begin{tabular}{ccc}
    \begin{minipage}[l]{0.33\hsize}
      \subfigure[GridWorld]{\includegraphics[width=\columnwidth]{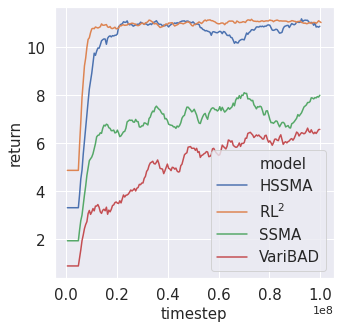}
      \label{fig:exp:lc:grid}
      }
    \end{minipage}
    \begin{minipage}[l]{0.33\hsize}
      \subfigure[HalfCheetahVel]{\includegraphics[width=\columnwidth]{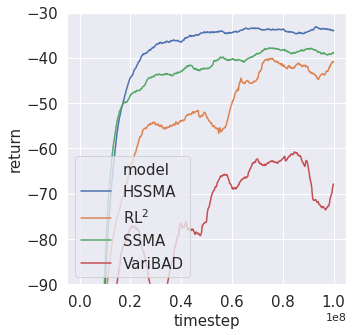}
      \label{fig:exp:lc:cheetah_vel}
      }
    \end{minipage}
    \begin{minipage}[l]{0.33\hsize}
      \subfigure[Walker]{\includegraphics[width=\columnwidth]{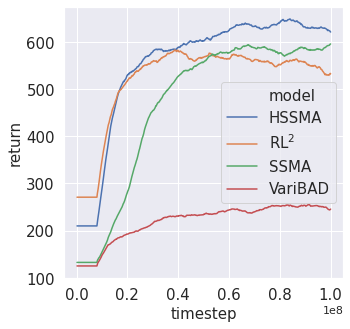}
      \label{fig:exp:lc:walker}
      }
    \end{minipage}
  \end{tabular}
  \caption{Learning curves for the (a) GridWorld, (b) HalfCheetahVel, and (c) Walker environments.
           The x-axis shows the number of frames used to train the policy whereas the y-axis shows the return (cumulative rewards) at the $N$-th episode ($N=4$ for (a) and $N=2$ for (b, c)).
  }
  \label{fig:exp:lc}
 \end{center}
 \vskip -0.4in
\end{figure}

Here we evaluated the training efficiency and the final meta-RL performance of the methods, as performed in \cite{rakelly2019pearl}.
The training efficiency is also significant because meta-RL policies typically require massive amounts of experience collected across a distribution of tasks, while the trained policies adapt to new tasks with only a few trials.
Figure \ref{fig:exp:lc} shows the learning curves of the returns at $N$-th episode.
It shows that, (i) RL$^2$ performed better (requires fewer training data and reaches higher final performance) than VariBAD in our {\em meta-POMDP} settings, although the opposite result had been reported in \cite{Zintgraf2020VariBAD} in their {\em meta-MDP} settings.
This indicates that considering the meta-POMDP structure is significant because the performance of VariBAD, which assumes the meta-MDP structure, drops in the meta-POMDP.
In addition, the figure shows that (ii) HSSMA achieved competitive or better results than the baseline methods.
Specifically, in GridWorld (\ref{fig:exp:lc:grid}), HSSMA has higher returns than SSMA and VariBAD, and competitive to RL$^2$.
Furthermore, in HalfCheetahVel (\ref{fig:exp:lc:cheetah_vel}) and Walker (\ref{fig:exp:lc:walker}), HSSMA has the highest return.
This is consistent with the past observations that RL$^2$ has difficulty in learning representation and decision making together as the observation and action space become larger.
In addition, HSSMA performed better than SSMA probably because the disentangled representation facilitates the policy training.

\section{Discussion and future work}

In this paper, we proposed estimation of the disentangled belief about the hidden task and states, and efficiently training a model-free policy with it.
In the experiments, we showed that the proposed method can learn the representation of task in the meta-POMDP environments, and that it outperformed the prior meta-RL methods.
Although we use the on-policy method (PPO) for fair comparison with the baseline methods, using the off-policy method (SAC) is an orthogonal approach that may further improve training efficiency of the policy.
Other future studies may reduce the gap between amortized distribution and posterior distribution by using normalizing flows, or apply the method to the semi-supervised situation where task information (such as language-instruction) is occasionally given by exploiting the disentangled representation of our model.

\acks{
  This work was supported by JSPS KAKENHI Grant Number JP20J11448.
}

\bibliography{ref}

\begin{thebibliography}{37}
\providecommand{\natexlab}[1]{#1}
\providecommand{\url}[1]{\texttt{#1}}
\expandafter\ifx\csname urlstyle\endcsname\relax
  \providecommand{\doi}[1]{doi: #1}\else
  \providecommand{\doi}{doi: \begingroup \urlstyle{rm}\Url}\fi

\bibitem[Alemi et~al.(2018)Alemi, Poole, Fischer, Dillon, Saurous, and
  Murphy]{alemi2018fixing}
Alexander Alemi, Ben Poole, Ian Fischer, Joshua Dillon, Rif~A Saurous, and
  Kevin Murphy.
\newblock Fixing a broken elbo.
\newblock In \emph{International Conference on Machine Learning}, pages
  159--168, 2018.

\bibitem[Bengio et~al.(2013)Bengio, Courville, and
  Vincent]{Bengio+2013_RLsurvey}
Y~Bengio, Aaron Courville, and Pascal Vincent.
\newblock Representation learning: A review and new perspectives.
\newblock \emph{IEEE transactions on pattern analysis and machine
  intelligence}, 35:\penalty0 1798--1828, 08 2013.
\newblock \doi{10.1109/TPAMI.2013.50}.

\bibitem[B{{\"o}}hmer et~al.(2013)B{{\"o}}hmer, Gr{{\"u}}new{{\"a}}lder, Shen,
  Musial, and Obermayer]{boehmer13sfa}
Wendelin B{{\"o}}hmer, Steffen Gr{{\"u}}new{{\"a}}lder, Yun Shen, Marek Musial,
  and Klaus Obermayer.
\newblock Construction of approximation spaces for reinforcement learning.
\newblock \emph{Journal of Machine Learning Research}, 14\penalty0
  (27):\penalty0 2067--2118, 2013.

\bibitem[Bowman et~al.(2016)Bowman, Vilnis, Vinyals, Dai, Jozefowicz, and
  Bengio]{bowman2016generating}
Samuel~R Bowman, Luke Vilnis, Oriol Vinyals, Andrew Dai, Rafal Jozefowicz, and
  Samy Bengio.
\newblock Generating sentences from a continuous space.
\newblock In \emph{Proceedings of The 20th SIGNLL Conference on Computational
  Natural Language Learning}, pages 10--21, 2016.

\bibitem[Chen et~al.(2017)Chen, Kingma, Salimans, Duan, Dhariwal, Schulman,
  Sutskever, and Abbeel]{chen2017vlae}
Xi~Chen, Diederik~P. Kingma, Tim Salimans, Yan Duan, Prafulla Dhariwal, John
  Schulman, Ilya Sutskever, and Pieter Abbeel.
\newblock {Variational Lossy Autoencoder}.
\newblock In \emph{Proc. 5th International Conference on Learning
  Representations}, 2017.

\bibitem[Chevalier-Boisvert et~al.(2019)Chevalier-Boisvert, Bahdanau, Lahlou,
  Willems, Saharia, Nguyen, and Bengio]{chevalier-boisvert2018babyai}
Maxime Chevalier-Boisvert, Dzmitry Bahdanau, Salem Lahlou, Lucas Willems,
  Chitwan Saharia, Thien~Huu Nguyen, and Yoshua Bengio.
\newblock Baby{AI}: First steps towards grounded language learning with a human
  in the loop.
\newblock In \emph{International Conference on Learning Representations}, 2019.

\bibitem[Duan et~al.(2016)Duan, Chen, Houthooft, Schulman, and
  Abbeel]{duan2016benchmarkRL}
Yan Duan, Xi~Chen, Rein Houthooft, John Schulman, and Pieter Abbeel.
\newblock Benchmarking deep reinforcement learning for continuous control.
\newblock volume~48 of \emph{Proceedings of Machine Learning Research}, pages
  1329--1338, New York, New York, USA, 20--22 Jun 2016. PMLR.

\bibitem[Duan et~al.(2017)Duan, Schulman, Chen, Bartlett, Sutskever, and
  Abbeel]{Duan2016RL2FR}
Yan Duan, John Schulman, Xi~Chen, Peter~L. Bartlett, Ilya Sutskever, and Pieter
  Abbeel.
\newblock Rl$^2$: Fast reinforcement learning via slow reinforcement learning.
\newblock In \emph{International Conference on Learning Representations}, 2017.

\bibitem[Finn et~al.(2017)Finn, Abbeel, and Levine]{finn2017maml}
Chelsea Finn, Pieter Abbeel, and Sergey Levine.
\newblock Model-agnostic meta-learning for fast adaptation of deep networks.
\newblock In Doina Precup and Yee~Whye Teh, editors, \emph{Proceedings of the
  34th International Conference on Machine Learning}, volume~70 of
  \emph{Proceedings of Machine Learning Research}, pages 1126--1135,
  International Convention Centre, Sydney, Australia, 06--11 Aug 2017. PMLR.

\bibitem[Finn et~al.(2018)Finn, Xu, and Levine]{finn2018pmaml}
Chelsea Finn, Kelvin Xu, and Sergey Levine.
\newblock Probabilistic model-agnostic meta-learning.
\newblock In S.~Bengio, H.~Wallach, H.~Larochelle, K.~Grauman, N.~Cesa-Bianchi,
  and R.~Garnett, editors, \emph{Advances in Neural Information Processing
  Systems 31}, pages 9516--9527. Curran Associates, Inc., 2018.

\bibitem[Gregor et~al.(2019)Gregor, Rezende, Besse, Wu, Merzic, and van~den
  Oord]{gregor2019shaping}
Karol Gregor, Danilo~Jimenez Rezende, Frederic Besse, Yan Wu, Hamza Merzic, and
  A{\"{a}}ron van~den Oord.
\newblock Shaping belief states with generative environment models for {RL}.
\newblock In \emph{Advances in Neural Information Processing Systems 32}, pages
  13475--13487, 2019.

\bibitem[Haarnoja et~al.(2018)Haarnoja, Zhou, Abbeel, and
  Levine]{haarnoja18sac}
Tuomas Haarnoja, Aurick Zhou, Pieter Abbeel, and Sergey Levine.
\newblock Soft actor-critic: Off-policy maximum entropy deep reinforcement
  learning with a stochastic actor.
\newblock volume~80 of \emph{Proceedings of Machine Learning Research}, pages
  1861--1870, Stockholmsm^^c3^^a4ssan, Stockholm Sweden, 10--15 Jul 2018. PMLR.

\bibitem[Han et~al.(2020)Han, Doya, and Tani]{han2020variational}
Dongqi Han, Kenji Doya, and Jun Tani.
\newblock Variational recurrent models for solving partially observable control
  tasks.
\newblock In \emph{International Conference on Learning Representations}, 2020.

\bibitem[Higgins et~al.(2017{\natexlab{a}})Higgins, Matthey, Pal, Burgess,
  Glorot, Botvinick, Mohamed, and Lerchner]{higgins2017beta}
Irina Higgins, Loic Matthey, Arka Pal, Christopher Burgess, Xavier Glorot,
  Matthew Botvinick, Shakir Mohamed, and Alexander Lerchner.
\newblock beta-vae: Learning basic visual concepts with a constrained
  variational framework.
\newblock In \emph{International Conference on Learning Representations},
  2017{\natexlab{a}}.

\bibitem[Higgins et~al.(2017{\natexlab{b}})Higgins, Pal, Rusu, Matthey,
  Burgess, Pritzel, Botvinick, Blundell, and Lerchner]{higgins2017darla}
Irina Higgins, Arka Pal, Andrei Rusu, Loic Matthey, Christopher Burgess,
  Alexander Pritzel, Matthew Botvinick, Charles Blundell, and Alexander
  Lerchner.
\newblock {DARLA}: Improving zero-shot transfer in reinforcement learning.
\newblock volume~70 of \emph{Proceedings of Machine Learning Research}, pages
  1480--1490, International Convention Centre, Sydney, Australia, 06--11 Aug
  2017{\natexlab{b}}. PMLR.

\bibitem[Hsieh et~al.(2018)Hsieh, Liu, Huang, Fei-Fei, and
  Niebles]{hsieh2018video}
Jun-Ting Hsieh, Bingbin Liu, De-An Huang, Li~F Fei-Fei, and Juan~Carlos
  Niebles.
\newblock Learning to decompose and disentangle representations for video
  prediction.
\newblock In S.~Bengio, H.~Wallach, H.~Larochelle, K.~Grauman, N.~Cesa-Bianchi,
  and R.~Garnett, editors, \emph{Advances in Neural Information Processing
  Systems 31}, pages 517--526. Curran Associates, Inc., 2018.

\bibitem[Hsu et~al.(2017)Hsu, Zhang, and Glass]{hsu2017fhvae}
Wei-Ning Hsu, Yu~Zhang, and James Glass.
\newblock Unsupervised learning of disentangled and interpretable
  representations from sequential data.
\newblock In \emph{Advances in Neural Information Processing Systems 30}, pages
  1878--1889. 2017.

\bibitem[Humplik et~al.(2019)Humplik, Galashov, Hasenclever, Ortega, Teh, and
  Heess]{humplik2019meta}
Jan Humplik, Alexandre Galashov, Leonard Hasenclever, Pedro~A. Ortega, Yee~Whye
  Teh, and Nicolas Heess.
\newblock Meta reinforcement learning as task inference.
\newblock \emph{CoRR}, abs/1905.06424, 2019.
\newblock URL \url{http://arxiv.org/abs/1905.06424}.

\bibitem[Igl et~al.(2018)Igl, Zintgraf, Le, Wood, and Whiteson]{igl2018dvrl}
Maximilian Igl, Luisa Zintgraf, Tuan~Anh Le, Frank Wood, and Shimon Whiteson.
\newblock Deep variational reinforcement learning for {POMDP}s.
\newblock volume~80 of \emph{Proceedings of Machine Learning Research}, pages
  2117--2126, Stockholmsm^^c3^^a4ssan, Stockholm Sweden, 10--15 Jul 2018. PMLR.
\newblock URL \url{http://proceedings.mlr.press/v80/igl18a.html}.

\bibitem[Jiang et~al.(2019)Jiang, Gu, Murphy, and Finn]{jiang2019lang}
YiDing Jiang, Shixiang~(Shane) Gu, Kevin~P Murphy, and Chelsea Finn.
\newblock Language as an abstraction for hierarchical deep reinforcement
  learning.
\newblock In \emph{Advances in Neural Information Processing Systems 32}, pages
  9419--9431. Curran Associates, Inc., 2019.

\bibitem[Jonschkowski and Brock(2015)]{Jonschkowski2015prior}
Rico Jonschkowski and Oliver Brock.
\newblock Learning state representations with robotic priors.
\newblock \emph{Auton. Robots}, 39\penalty0 (3):\penalty0 407^^e2^^80^^93428,
  October 2015.
\newblock ISSN 0929-5593.
\newblock \doi{10.1007/s10514-015-9459-7}.

\bibitem[Kaelbling et~al.(1998)Kaelbling, Littman, and
  Cassandra]{leslie1998pomdp}
Leslie~Pack Kaelbling, Michael~L. Littman, and Anthony~R. Cassandra.
\newblock Planning and acting in partially observable stochastic domains.
\newblock \emph{Artif. Intell.}, 101\penalty0 (1^^e2^^80^^932):\penalty0
  99^^e2^^80^^93134, May 1998.
\newblock ISSN 0004-3702.

\bibitem[Kingma and Welling(2014)]{kingma2013auto}
Diederik~P Kingma and Max Welling.
\newblock Auto-encoding variational bayes.
\newblock In \emph{International Conference on Learning Representations}, 2014.

\bibitem[Kingma et~al.(2014)Kingma, Mohamed, Jimenez~Rezende, and
  Welling]{kingma2014ssvae}
Durk~P Kingma, Shakir Mohamed, Danilo Jimenez~Rezende, and Max Welling.
\newblock Semi-supervised learning with deep generative models.
\newblock In \emph{Advances in Neural Information Processing Systems 27}, pages
  3581--3589. 2014.

\bibitem[Kingma et~al.(2016)Kingma, Salimans, Jozefowicz, Chen, Sutskever, and
  Welling]{kingma2016iaf}
Durk~P Kingma, Tim Salimans, Rafal Jozefowicz, Xi~Chen, Ilya Sutskever, and Max
  Welling.
\newblock Improved variational inference with inverse autoregressive flow.
\newblock In D.~D. Lee, M.~Sugiyama, U.~V. Luxburg, I.~Guyon, and R.~Garnett,
  editors, \emph{Advances in Neural Information Processing Systems 29}, pages
  4743--4751. Curran Associates, Inc., 2016.

\bibitem[Lee et~al.(2019)Lee, Nagabandi, Abbeel, and Levine]{lee2019slac}
Alex~X. Lee, Anusha Nagabandi, Pieter Abbeel, and Sergey Levine.
\newblock Stochastic latent actor-critic: Deep reinforcement learning with a
  latent variable model.
\newblock \emph{arXiv preprint arXiv:1907.00953}, 2019.

\bibitem[Lesort et~al.(2018)Lesort, Rodr{\'i}guez, Goudou, and
  Filliat]{Lesort2018StateRL}
Timoth{\'e}e Lesort, Natalia~D{\'i}az Rodr{\'i}guez, Jean-Fran^^c3^^a7ois
  Goudou, and David Filliat.
\newblock State representation learning for control: An overview.
\newblock \emph{Neural networks : the official journal of the International
  Neural Network Society}, 108:\penalty0 379--392, 2018.

\bibitem[Maal\o~e et~al.(2019)Maal\o~e, Fraccaro, Li\'{e}vin, and
  Winther]{maalo2019biva}
Lars Maal\o~e, Marco Fraccaro, Valentin Li\'{e}vin, and Ole Winther.
\newblock Biva: A very deep hierarchy of latent variables for generative
  modeling.
\newblock In \emph{Advances in Neural Information Processing Systems 32}, pages
  6548--6558. 2019.

\bibitem[Mnih et~al.(2015)Mnih, Kavukcuoglu, Silver, Rusu, Veness, Bellemare,
  Graves, Riedmiller, Fidjeland, Ostrovski, Petersen, Beattie, Sadik,
  Antonoglou, King, Kumaran, Wierstra, Legg, and Hassabis]{mnih2015humanlevel}
Volodymyr Mnih, Koray Kavukcuoglu, David Silver, Andrei~A. Rusu, Joel Veness,
  Marc~G. Bellemare, Alex Graves, Martin Riedmiller, Andreas~K. Fidjeland,
  Georg Ostrovski, Stig Petersen, Charles Beattie, Amir Sadik, Ioannis
  Antonoglou, Helen King, Dharshan Kumaran, Daan Wierstra, Shane Legg, and
  Demis Hassabis.
\newblock Human-level control through deep reinforcement learning.
\newblock \emph{Nature}, 518\penalty0 (7540):\penalty0 529--533, February 2015.
\newblock ISSN 00280836.

\bibitem[Rakelly et~al.(2019)Rakelly, Zhou, Finn, Levine, and
  Quillen]{rakelly2019pearl}
Kate Rakelly, Aurick Zhou, Chelsea Finn, Sergey Levine, and Deirdre Quillen.
\newblock Efficient off-policy meta-reinforcement learning via probabilistic
  context variables.
\newblock In Kamalika Chaudhuri and Ruslan Salakhutdinov, editors,
  \emph{Proceedings of the 36th International Conference on Machine Learning},
  volume~97 of \emph{Proceedings of Machine Learning Research}, pages
  5331--5340, Long Beach, California, USA, 09--15 Jun 2019. PMLR.

\bibitem[Ross et~al.(2008)Ross, Chaib-draa, and Pineau]{stephane2008bapomdp}
Stephane Ross, Brahim Chaib-draa, and Joelle Pineau.
\newblock Bayes-adaptive pomdps.
\newblock In J.~C. Platt, D.~Koller, Y.~Singer, and S.~T. Roweis, editors,
  \emph{Advances in Neural Information Processing Systems 20}, pages
  1225--1232. Curran Associates, Inc., 2008.

\bibitem[Schulman et~al.(2017)Schulman, Wolski, Dhariwal, Radford, and
  Klimov]{schulman2017ppo}
John Schulman, Filip Wolski, Prafulla Dhariwal, Alec Radford, and Oleg Klimov.
\newblock Proximal policy optimization algorithms.
\newblock \emph{CoRR}, abs/1707.06347, 2017.
\newblock URL \url{http://arxiv.org/abs/1707.06347}.

\bibitem[{Todorov} et~al.(2012){Todorov}, {Erez}, and
  {Tassa}]{todorov2012mujoco}
E.~{Todorov}, T.~{Erez}, and Y.~{Tassa}.
\newblock Mujoco: A physics engine for model-based control.
\newblock In \emph{2012 IEEE/RSJ International Conference on Intelligent Robots
  and Systems}, pages 5026--5033, 2012.
\newblock \doi{10.1109/IROS.2012.6386109}.

\bibitem[Vahdat and Kautz(2020)]{vahdat2020NVAE}
Arash Vahdat and Jan Kautz.
\newblock {NVAE}: A deep hierarchical variational autoencoder.
\newblock In \emph{Neural Information Processing Systems (NeurIPS)}, 2020.

\bibitem[Yingzhen and Mandt(2018)]{yingzhen2018dsae}
Li~Yingzhen and Stephan Mandt.
\newblock Disentangled sequential autoencoder.
\newblock In \emph{Proceedings of the 35th International Conference on Machine
  Learning}, pages 5670--5679, 2018.

\bibitem[Zhao et~al.(2020)Zhao, Nagabandi, Rakelly, Finn, and
  Levine]{zhao2020meld}
Tony~Z. Zhao, Anusha Nagabandi, Kate Rakelly, Chelsea Finn, and Sergey Levine.
\newblock Meld: Meta-reinforcement learning from images via latent state
  models.
\newblock 2020.

\bibitem[Zintgraf et~al.(2020)Zintgraf, Shiarlis, Igl, Schulze, Gal, Hofmann,
  and Whiteson]{Zintgraf2020VariBAD}
Luisa Zintgraf, Kyriacos Shiarlis, Maximilian Igl, Sebastian Schulze, Yarin
  Gal, Katja Hofmann, and Shimon Whiteson.
\newblock Varibad: A very good method for bayes-adaptive deep rl via
  meta-learning.
\newblock In \emph{International Conference on Learning Representations}, 2020.

\end{thebibliography}

\end{document}